\documentclass{bmvc2k}
\usepackage{epstopdf}

\title{Frustratingly Easy Person Re-Identification: Generalizing Person Re-ID in Practice}

\addauthor{Jieru Jia}{12112059@bjtu.edu.cn}{1}
\addauthor{Qiuqi Ruan}{qqruan@bjtu.edu.cn}{1}
\addauthor{Timothy M. Hospedales}{t.hospedales@ed.ac.uk}{2}

\addinstitution{
	Institute of Information Science\\
	 Beijing Jiaotong University\\
	  Beijing, China}
\addinstitution{
 Institute of Perception, Action and
 Behaviour\\
 University of Edinburgh,\\
 Edinburgh, UK
}

\runninghead{JIA ET Al.}{Generalizing Person Re-ID in Practice}

\def\eg{\emph{e.g}\bmvaOneDot}

\def\etal{\emph{et al}\bmvaOneDot}
\newcommand{\keypoint}[1]{\vspace{0.1cm}\noindent\textbf{#1}\quad}
\newcommand{\reid}{Re-ID}

\newcommand{\name}{DualNorm}
\newcommand{\cut}[1]{}

\begin{document}	
	\maketitle	
	\begin{abstract}
		Contemporary person  re-identification (\reid) methods usually require access to data from the deployment camera network during training in order to perform well. This is because contemporary \reid{} models trained on one dataset do not generalise to other camera networks due to the domain-shift between datasets. This requirement is often the bottleneck for deploying \reid{} systems in practical security or commercial applications, as it may be impossible to collect this data in advance or prohibitively costly to annotate it. This paper alleviates this issue by proposing a simple baseline for  domain generalizable~(DG) person re-identification. That is, to learn a \reid{} model from a set of source domains that is suitable for application to unseen datasets out-of-the-box, without any model updating. Specifically, we observe that the domain discrepancy in \reid{} is due to style and content variance across datasets and demonstrate appropriate Instance and Feature Normalization alleviates much of the resulting domain-shift in Deep \reid{} models.  Instance Normalization~(IN) in early layers filters out style statistic variations and Feature Normalization~(FN) in deep layers is able to further eliminate disparity in content statistics. Compared to contemporary alternatives, this approach is extremely simple to implement, while being faster to train and test, thus making it an extremely valuable baseline for implementing \reid{} in practice. With a few lines of code, it increases the rank 1 \reid{} accuracy by {11.8\%, 33.2\%, 12.8\% and 8.5\%} on the VIPeR, PRID, GRID, and i-LIDS benchmarks respectively. Source codes are available at \url{https://github.com/BJTUJia/person_reID_DualNorm}.
	\end{abstract}
	
	\section{Introduction}
	\label{sec:intro}
	Person re-identification (\reid{}) aims to match pedestrian images captured by different cameras at different times and locations. Despite their goal of cross-camera matching, most contemporary \reid{} methods overfit to specific datasets (camera networks) in that, once trained on a given dataset, they perform poorly if applied to a different camera network. This prevents a single \reid{} system from being built that can successfully apply off-the-shelf to diverse scenarios. Contemporary research focuses on supervised \reid{}, where systems are trained with annotated data from the deployment network \cite{cheng2016person,zhao2017spindle} -- at a significant cost to scalability; or unsupervised domain adaptation (UDA) \reid{} \cite{deng2018image,2018eanet,zhong2019invariance,wang2018transferable}, which aims to alleviate the annotation cost by adapting a model trained on annotated source datasets (whose annotation cost is then amortised) to unannotated target datasets. While UDA methods reduce annotation costs, they still require data collection for the target network prior to model adaptive training. Since it may not be known in advance where a \reid{} model should be deployed, these are not ideal for real-world applications.
	
	In this paper, we consider the most practically valuable problem setting of domain generalization~(DG) \reid{}, in which we train using only labeled source dataset images, without touching any target network images -- either labeled or unlabelled. During testing the model is applied to novel unseen datasets with no adaptation.  This setting simulates the ideal scenario in which a strong learner is trained once in the lab, and can then be deployed to diverse camera networks in the wild with no further data collection or adaptive training required. This DG setting is the most practically relevant for real-world commercial/security applications. However due to its challenging nature, few contemporary deep-learning-based methods have attempted the DG setting, besides the recent DIMN \cite{dimn}. DIMM is effective but requires a complicated meta-learning procedure for training and dynamic model-synthesis at testing-time that makes it relatively slow and cumbersome for practical applications. An easy-to-implement, fast, and stable DG \reid{} method is still needed in practice.
	
	Our goal is to train a model that generalizes well to unseen domains out-of-the-box -- recognising new identities in a novel camera network. Our \emph{\name} solution explicitly addresses domain biases via normalization in deep feature extraction. We observe that the discrepancy between diverse \reid{} domains arises from two sources: (i)~difference in style, \eg, distinct color saturations and contrasts, lighting, resolutions, clothing styles, seasons; and (ii)~content variation caused by differing typical distance to objects, focal length, viewpoint and typical poses. To alleviate these issues, we design a CNN architecture that jointly normalizes style and content statistics. Specifically, we use Instance Normalization (IN) layers at each bottleneck in shallow layers to capture and eliminate  style variations. We next normalise different content statistics by performing Batch Normalization (BN)-based normalisation of extracted features. By jointly tackling style and content variances to explicitly address domain bias, extracted features are more universal and domain agnostic.
	
	Our method is extremely simple compared to elaborate contemporary alternatives~\cite{dimn,2018eanet,zhong2019invariance}. However, we argue that this should be considered a plus in practice where ease of implementation, stability, and efficiency are more valuable than elaborate methodologies. Furthermore, we point to a distinguished history of methods in vision and pattern recognition, where simple baselines that surpassed their more elaborate predecessors made significant impact \cite{daume2007easyDA,paredes2015ez_zsl,jabri2016revisitVQA}.
	
	To summarize, the main contributions of this paper are: (1)~A strong baseline for domain generalization person \reid{}. Our method requires no information about the target domain during training, and works well as an out-of-the-box feature extractor for novel camera networks. It surpasses most existing (target domain)-supervised methods, and all existing methods that do not learn on target data. (2)~We present a simple yet highly efficient way to address domain-shift through normalization layers: IN in shallow layers and BN in deep layers alleviate style and content statistic domain biases. (3)~Our approach is very easy to implement (with just a few lines of code) and fast to run, but boosts DG \reid{} performance by a large margin {for both MobileNet and ResNet backbones}. Its simplicity, efficacy and efficiency make it appealing for \reid{} in practice.
	
	\section{Related Work}
	
	\keypoint{Domain adaption and generalization}
	Unsupervised Domain Adaptation~(UDA) alleviates domain-shift without recourse to \emph{annotated} data in the target domain \cite{chang2019dlstl}. For example, by reducing the Maximum Mean Discrepancy (MMD) \cite{gretton2007kernel} between domains \cite{yan2017mind}, or training an adversarial domain-classifier \cite{tzeng2017adversarial} to make different domains indistinguishable. In the \reid{} community, the UDA methods typically resort to image-synthesis \cite{SyRI,deng2018image} or focus on source-target domain alignment \cite{zhong2019invariance,2018eanet,wang2018transferable}. While these methods are annotation efficient, they do require prior collection of target-domain data for training, while our method has no such requirement, making it more valuable in practice where the deployment network is not known at the time of model creation.
	
	Compared to UDA\cite{deng2018image,2018eanet}, Domain Generalisation (DG) methods \cite{li2019featureCritic} aim to create models that are robust-by-design to domain-shift between training and testing. These methods tend to leverage architectures specially designed for domain-shift robustness \cite{khosla2012undoingDatasetBias,ghifary2015gdMTLae}, or propose meta-learning procedures for standard architectures \cite{balaji2018metaReg,li2018mldg,li2019featureCritic}. Our method is in the former category, but only requires a minor modification of standard \reid{} architectures.  In a \reid{} context, we are only aware of DIMN \cite{dimn} as a contemporary attempt at the DG problem setting, which uses a meta-learning approach\footnote{Of course classic feature-engineering approaches \cite{farenzena2010reidentify_symmetry} are not tied to specific datasets, but these are not competitive with contemporary deep-learning based approaches.}. While DIMN is effective, it requires a complicated and slow meta-learning procedure for training, which limits its appeal to practitioners. Furthermore, DIMN uses dynamic model synthesis at runtime so it is not amenable to modifications for runtime scalability such as binarization, approximate nearest-neighbour search, and hashing. In contrast our carefully designed feature extractor is faster out-of-the-box, and can potentially be extended in all of these ways.
	
	\keypoint{Normalization}
	Batch Normalization (BN) \cite{ioffe2015batch} has become a key technique in CNN training, by standardizing input data or activations using  statistics computed over  examples in a mini-batch. Instance Normalization (IN) \cite{ulyanov2016instance} performs BN-like computation over a single sample. Moreover, the IBN building block recently proposed in \cite{pan2018two} enhances models' generalization ability by integrating IN and BN. A different way to combine BN and IN was put forward in \cite{nam2018batch}. Recently, some effort has been made in feature normalization \cite{xu2018unsupervised,zheng2018ring},  mainly applying ${l_2}$-norm to the feature embeddings, constraining them to the unit circle.
	
	\keypoint{Difference from previous works}	
	{The normalization techniques we explore have been largely presented in existing works. The contribution of our work lies in the following aspects: (1)~We explore the feasibility of IN in inverted residual bottlenecks in MobileNetV2 \cite{Sandler2018MobileNetV2} to overcome the style variances in \reid{}. (2)~Although BN layers after feature extraction are widely used, we explicitly utilize them as a means of feature normalization for domain-shift robustness. (3)~We systematically demonstrate that, by integrating the two normalization techniques into a unified network with end-to-end learning,  our  {\name} provides the strongest baseline for person \reid{}, while being simple and requiring no target domain data.}

	\section{Method}
	\keypoint{Setup} For domain generalization person re-ID, we assume $K$ labeled source datasets $D=\left\{ {{D_1},{D_2}, \ldots {D_K}} \right\}$. Each source ${D_i} = \left\{ {{X^{\left( i \right)}},{Y^{\left( i \right)}}} \right\}$ consists of image-label pairs, where ${y_i} \in \left\{ {1,2, \ldots {M_i}} \right\}$ and ${{M_i}}$ is the number of identities in ${D_i}$ \cite{dimn}. Since the label spaces for $K$ source sets are disjoint, we use their union as label space with $N = \sum\limits_i {{M_i}} $ identities in total.
	
	\keypoint{A Strong Cross-domain Baseline} Following \cite{dimn}, we build a strong starting model by aggregating the labeled images from multiple source domains, and training a single model to discriminate all $N$ identities. During the testing phase, the learnt model is used as an off-the-shelf feature extractor on previously unseen identities. We choose lightweight MobileNetV2 \cite{Sandler2018MobileNetV2} as backbone, since it has significantly fewer parameters compared to other CNN architectures like ResNet \cite{He2016Deep}. We keep the default structure of MobileNetV2 except for changing the dimension of last classification layer (FC) to be $N$, i.e., the total number of identities. We use the cross-entropy~(CE) loss to calculate the loss of all source domains:
	\begin{equation}
	{L_{CE}} = - \frac{1}{{{n_s}}}\sum\limits_{i = 1}^{{n_s}} {\log p\left( {{y_i}|{x_i}} \right)}
	\end{equation}
	where ${n_s}$ is the number of images in a mini-batch and ${p\left( {{y_i}|{x_i}} \right)}$ is the predicted probability that the source image ${{x_i}}$ belongs to identity  
	${{y_i}}$. It is worth mentioning the mini-batch scheduling: each mini-batch contains samples randomly selected from all source domains. The model is trained from scratch, without pre-training on ImageNet \cite{deng2009imagenet}.
	
	\keypoint{Testing} During testing phase, given an input image from target domain, we extract the 1280-d pooling layer activations as output features. Then, we use Euclidean distance to perform person retrieval in the target set. Next, we will discuss how to improve this na\"ive approach to achieve better generalization through normalizing dataset-specific biases.			
	\subsection{Style Normalization}
	\begin{figure}[t]
		\centering	
		\includegraphics[width=5.2cm,height=4.4cm]{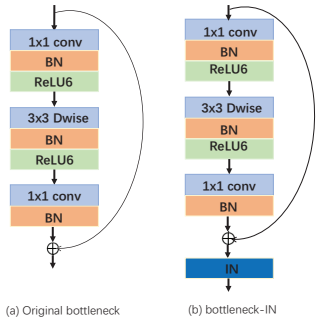}	
		\caption{The structure of our MobileNetV2 bottlenecks with added Instance normalization (we use the block with stride=1 as an example).}
		\label{fig1}
	\end{figure}
	The structure of our bottleneck with IN layer is shown in Fig.~\ref{fig1}. In modern deep networks, Batch Normalization (BN) \cite{ioffe2015batch} has become a standard operator with its ability to stabilize and accelerate training. On the other hand, Instance Normalization (IN) \cite{ulyanov2016instance}, which is mainly used in style transfer tasks \cite{nam2018batch,pan2018two}, normalizes feature responses at the instance-level with the statistics of a single sample. It is assumed that the styles of images are encoded in the first-order statistics, i.e. the mean and variance of a convolutional feature map \cite{nam2018batch}. Therefore, by normalizing the output of the original inverted residual bottlenecks, we filter out instance specific style variances, which makes the learned features more dataset-agnostic. 
	
	According to the findings in \cite{pan2018two}, the appearance divergence caused by style variances mostly lies in early layers and adding IN in early layers can effectively reduce the domain-shift due to style differences. On the other hand, applying IN to deep layers may cause severe loss of discriminative information and degrade the classification performance. Therefore, we only add IN for the shallow layers of the network.

\subsection{Content Normalization}
\begin{figure}[t]
		\centering			
		\includegraphics[width=1.0\textwidth]{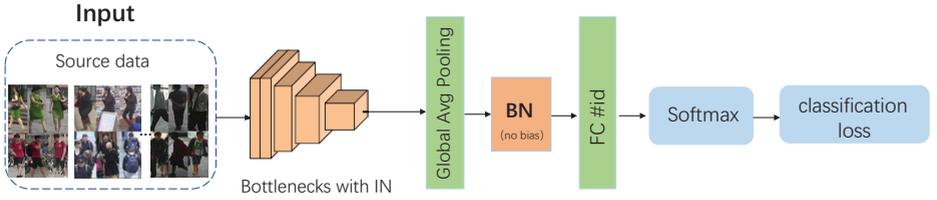}		
		\caption{Structure of our \name{} network. Input images are randomly sampled from multiple source sets, and go forward through the stacked bottlenecks with IN. After global average pooling, a BN with bias removed is adopted to normalize the features. They are then fed into a classifier with FC and softmax layer.}
		\label{fig2}
	\end{figure}

	\keypoint{Summary} The feature norm methods in \cite{xu2018unsupervised,zheng2018ring} help feature robustness by introducing smoothness with ${l_2}$-norm, which is also the fundamental effect of BN, as explained in \cite{santurkar2018does}. Hence, we exploit BN as an alternative feature norm approach. The exploitation of BN in our overall framework is shown in Fig.~\ref{fig2}. In this case,  BN is applied to the penultimate layer after feature encoding.
	
	\keypoint{Details} 
	Let $F$ and $G$ denote the feature extraction module and classifier layer respectively. The input images ${\left\{ {{x_n}} \right\}_{n = 1, \ldots N}}$ are passed through $F$ to acquire the unnormalized feature $F\left( {x_n} \right) \in {R^{B \times C \times H \times W}}$, where $H$ and $W$ indicate spatial location, $C$ is the number of channels and $B$ denotes the number of examples in the mini-batch. With a global average pooling layer, $F\left( {x_n} \right)$ is transformed to $f\left( {x_n} \right) \in {R^{B \times C \times 1 \times 1}}$. Next, BN normalizes each channel of $f\left( {x_n} \right)$ using the mean $\mu$ and variance $\sigma^2$ computed over the mini-batch: 
	\begin{equation}
	\setlength{\abovedisplayskip}{0pt}
	\setlength{\belowdisplayskip}{3pt}
	BN\left( {f\left( {{x_n}} \right)} \right) = \gamma \left( {\frac{{f\left( {{x_n}} \right) - \mu }}{\sigma }} \right) + \beta
	\end{equation}
	This causes the feature for a particular instance to depend on other instances in the mini-batch and helps regularize the network training. Another advantage of this BatchNorm is that it makes gradients more stable and enables larger learning rates, thus yielding faster convergence and better generalization.
	
	Note that different to vanilla BN, we fix the bias to $\beta=0$ following \cite{luo2019bags}. This is because the softmax loss is angular and the hypersphere is nearly symmetric about the origin of the coordinate axis. By removing the bias parameter, the features are constrained around the origin of the coordinate axis. Accordingly, the bias of the following FC layer is also removed.
	
	\keypoint{Discussion} 
	By using the proposed {\name} mechanism, we are able to push the learned feature space towards domain-invariant and better generalizable to novel domains. More specifically, we assume that the domain discrepancy in \reid{} mainly stems from difference in image style and content. IN in early layers can explicitly normalizes domain-specific image styles, like color, lighting, clothing style \etal. BN after feature extractor is able to normalize the disparity in content statistics caused by differing camera angle, body size, viewpoint, pose \etal. Using the style and content normalisations outlined above, our feature extractor alleviates much of the domain-specific biases that cause the domain-shift problem when porting re-id models across datasets. Therefore, the learned model is more likely to be domain agnostic and own better generalization ability.
		
	The added normalization layers only introduce negligible additional parameters, taking up a very small percentage of full network parameters. The additional parameters are learned jointly with the other parameters of the network in an end-to-end manner, requiring very limited extra computation cost and GPU memory. Despite its simplicity, our method achieves the new state-of-the-art cross-domain re-ID accuracy. Thus it makes a major contribution towards making person \reid{} easy to implement and deploy.
	
	\section{Experiments}
	\subsection{Datasets and Settings}
	\keypoint{Datasets}
	To evaluate the proposed method, we follow \cite{dimn} and combine existing large-scale \reid{} datasets to form the source domains and test the performance on several smaller target datasets. This is to reflect the most desirable practical scenario where a practitioner would use the largest available existing datasets to train a deep \reid{} model, and hope that it can be applied off-the-shelf to a novel camera network. To be specific, we exploit Market-1501 \cite{zheng2015scalable}, DukeMTMC-reID \cite{zheng2017unlabeled}, CUHK02 \cite{li2013locally}, CUHK03 \cite{LiZhao-56} and 
	PersonSearch \cite{Tong2016End} as the source
	datasets, with a a total of $18,530$ training identities and 121,765 training images in all. We denote this large scale re-ID dataset collection as `MS'~(multi-source).
		
	The target datasets include VIPeR\cite{GrayTao-24}, PRID\cite{HirzerBeleznai-58}, QMUL GRID \cite{loymulti} and i-LIDS \cite{ZhengGong-23}. For the convenience of comparing with prior reported results, we use the standard evaluation protocols on the testing datasets. More specifically, we follow the single-shot setting with the number of probe/gallery images set as: VIPeR: 316/316; PRID: 100/649; GRID: 125/900; i-LIDS: 60/60 respectively. Note that for i-LIDS, two images per identity are randomly selected as gallery and probe respectively, following the settings in \cite{dimn}. For all the testing datasets, the average results over 10 random splits are reported. 
	
	\keypoint{Implementation Details}
	We use MobileNetV2 \cite{Sandler2018MobileNetV2} with width multiplier of 1.0 as the backbone network. In terms of input images, we keep the aspect ratio and resize them to $256 \times 128$. Random cropping and random flipping are applied as data augmentation. We implement our model with PyTorch \cite{paszke2017automatic} and train it on a single Titan X GPU. The model is trained from scratch with initial learning rate as 0.01 for all the layers. The SGD optimizer is exploited in a total of 150 epochs and the learning rate is divided by 10 after 100 epochs. The mini-batch size is set to 64. The only trick we use is label smoothing \cite{zheng2017unlabeled,luo2019bags}, which is  helpful to prevent the model from overfitting to  training IDs.
	
	\keypoint{Competitors} We compare our proposed \name{} method with a variety of alternatives. \textbf{`DG' Methods:} In terms of domain generalisation approaches, we compare with the domain aggregation baseline \textbf{AGG} (our method prior to adding normalisation layers) and state-of-the-art method DIMN \cite{dimn}. \textbf{`U' Methods:} For unsupervised methods, we compare with UDA methods MMFA \cite{Shan2018Multi}, TJ-AIDL \cite{wang2018transferable}, SyRI\cite{SyRI}, and a representative selection of other unsupervised alternatives. We also compare a representative selection of recent supervised (\textbf{`S'}) methods -- that use target images and labels. It is important to note that the UDA, and S methods are not fair competitors in that they use more information about the target domain than ours. We include them not as direct competitors, but to contextualise our results.
	
	\subsection{Comparisons Against State-of-the-Art}

	\begin{table}[t]
		\centering
		\begin{tabular}{l|c|c|l|l|l|l}
			\hline
			Method & Type & Source &VIPeR & PRID & GRID & i-LIDS  \\		
			\hline
			MMFA \cite{Shan2018Multi} & U  & Market &39.1 &35.1&-&-\\
			MMFA \cite{Shan2018Multi} & U  & Duke &36.3 &34.5&-&-\\
			TJ-AIDL\cite{wang2018transferable} & U & Market& 38.5 &26.8&-&-\\
			TJ-AIDL\cite{wang2018transferable} & U & Duke& 35.1 &34.8&-&-\\
			SyRI\cite{SyRI} & U& R+S\footnotemark[1]& 43.0& 43.0&-&56.5 \\
			CAMEL \cite{yu2017cross} & U & JSTL \cite{xiao2016learning} & 30.9& - &-&-\\			
			UMDL\cite{peng2016unsupervised} & U & Comb \footnotemark[2] & 31.5 &24.2&-&49.3 \\
			SSDAL \cite{su2016deep} & U & PETA \cite{deng2014pedestrian} & 37.9 &20.1&19.1&-\\
			\hline
			OneShot \cite{bak2017one} & S & Target & 34.3&41.4&-&51.2\\
			NFST \cite{ZhangXiang-28} & S & Target & 51.2&40.9&-&-\\
			Ensembles \cite{paisitk}& S & Target & 45.9&17.9&-&50.3\\
			DSPSL \cite{li2018discriminative}  & S & Target &-&-&-& 55.2 \\
			MTDnet \cite{chen2017multi} & S & Target & 47.5&32.0&-&58.4\\
			ImpTrpLoss \cite{cheng2016person} & S & Target & 47.8&22.0&-&60.4\\
			GOG+XQDA \cite{MatsukawaOkabe-9} & S & Target & 49.7 & \textbf{68.4}&24.7&-\\
			JLML \cite{li2017person} & S & Target&  50.2 & -&37.5&-\\
			SSM \cite{bai2017scalable} & S & Target & 53.7 &-&27.2&-\\
			SpindleNet \cite{zhao2017spindle} & S & Target & 53.8 &67.0&-&66.3\\
			\hline			
			AGG~(DIMN)\cite{dimn} & DG & MS &42.9 &38.9&29.7&69.2\\
			AGG~(Ours) & DG & MS &42.1 &27.2&28.6&66.3\\
			DIMN \cite{dimn} & DG & MS& 51.2 &	39.2&29.3&70.2\\
			\name (Ours) & DG & MS & \textbf{53.9}& 60.4&\textbf{41.4}&\textbf{74.8}\\
			\hline
		\end{tabular}
		\caption{Comparison against state-of-the-art on VIPeR, PRID, GRID and i-LIDS (Rank 1 accuracy). `U' and `S' denote (U)nsupervised and (S)upervised use of the target dataset for training. `DG' methods do not touch the target during training. `-': unreported result. }
		\label{tableviper}
	\end{table}
	\footnotetext[1] {`R' include CUHK03+Duke while `S' means Synthetic images.}
	\footnotetext[2] {Four out of five datasets VIPeR, PRID, CUHK01, i-LIDS and CAVIAR are combined as the source datasets when the fifth one is chosen as the target.}
	
		The rank 1 accuracy on VIPeR, PRID, GRID and i-LIDS are listed in Table~\ref{tableviper}. Models in the Unsupervised (U) setting are trained with some source data and then adapted to the target datasets using unlabeled images form their training split, while models in the Supervised (S) setting are trained with data and labels from the target dataset's training split. From Table~\ref{tableviper}, the following observations can be made: (1) The na\"ive aggregation solution provides a strong baseline. The use of a large amount of source data provides a stronger baseline than most existing unsupervised adaptation methods, even though the target data is not used at all. (2) Our approach significantly and consistently outperforms most Unsupervised and Supervised competitors, despite using neither images nor labels from the target domain. The ability to outperform recent fully-supervised approaches such as SpindleNet is noteworthy since we use neither the target data nor labels that they use. This shows that sufficient source data volume, and appropriate normalisation for debiasing can eventually alleviate the need for target-domain data in achieving state-of-the-art performance. This is a crucially important practical result, because it means that a model can be trained once and then re-deployed to different locations without re-training -- which is crucial for making \reid{} deployment practical. (3) Our approach significantly and consistently outperforms DIMN \cite{dimn}, the only purpose-designed DG competitor. This is despite the fact that our method is significantly simpler and more computationally efficient. 
		
	\subsection{Additional Analysis} 
	\keypoint{The effect of IN and FN}
	There are two important components in our framework: IN in early bottlenecks to alleviate style variance and FN on the penultimate layer to normalize content. To evaluate the contribution of each component, we separately add IN and FN to the backbone network and compare the performance in Table~\ref{ab0}. We can see that both IN and FN contribute notably to the improvement of overall performance. Among them, adding IN in early layers seems more beneficial, and combining two normalization techniques leads to a further performance gain, proving that they are complementary. Overall, our approach boosts the rank 1 rates of \textbf{AGG} by 11.8\%, 33.2\%, 12.8\%, 8.5\% respectively, only with extra normalization layers that introduce few paremeters.
	\begin{table}[t]
		\centering
		\begin{tabular}{l|c|c|l|l}
			\hline
			Components & VIPeR & PRID & GRID & i-LIDS \\
			\hline
			Baseline &  42.1 & 27.2 & 28.6 & 66.3\\				
			+IN & 52.7 & 54.1 & 39.7 & 70.5\\
			+FN & 48.1 & 47.2 & 31.3 & 71.2\\
			\name & 53.9 & 60.4 & 41.4 & 74.8\\
			\hline
		\end{tabular}		
		\caption{Ablation study on the impact of different components for cross-domain \reid{}.}
		\label{ab0}
	\end{table}
	
	\keypoint{On the location and amount of IN} 
	We apply FN at one fixed location (after feature pooling), while the best place to apply IN is unclear. We thus investigate where to apply IN. We denote the initial convolution block in MobileNetV2 \cite{Sandler2018MobileNetV2} as `Conv1'. The following groups of inverted residual bottlenecks are referred to as `Conv2\_x-Conv8\_x' respectively. 
	
	Table~\ref{ab1} gives performance of IN layers added to various locations in the original MobileNetV2. Note that for all the experiments here, the feature norm is removed to shed light on IN's influence. It can be seen that:~(1) adding more IN layers in early layers increases the performance more, but IN applied to deep layers has a detrimental effect. This indicates that a moderate amount of IN operations in shallow layers suffices. (2) Comparing `1-6' and `2-6', we can see the instance normalization on Conv1 is critical since it directly takes the activation of the first convolution layer as input, where  appearance differences are intrinsic.
	
	\begin{table}[t]
		\centering
		\begin{tabular}{l|c|c|l|l}
			\hline
			Groups & VIPeR & PRID & GRID & i-LIDS \\
			\hline
			None &  42.1 & 27.2 & 28.6 & 66.3\\
			1-3  & 43.9 & 37.2 & 29.4 &  67.1\\			
			1-4  & 48.0 & 42.3 & 31.9 & 69.4\\
			1-5  & 49.8 & 47.4 & 37.1 & 69.7\\			 
			1-6 & 52.7 & 54.1 & 39.7 & 70.5\\
			2-6 & 48.2 & 45.1 & 30.9 & 67.1\\
			1-7 & 46.3 & 42.9 & 34.8 & 66.0\\
			1-8 & 42.3 & 37.3 & 29.2 & 65.1\\
			\hline
		\end{tabular}		
		\caption{Design-space analysis on where to add IN layers in MobileNetV2 convolution blocks (rank 1 accuracy).}
		\label{ab1}
	\end{table}
	
	\keypoint{Comparison with other feature norm approaches}
	To investigate the effect of the added BN as feature normalization, we compare it with other recently proposed feature norm methods in Table~\ref{ab6}. Here, IN is added to Conv1-6 for all the experiments and we follow the default parameter settings in each paper for the compared methods. We observe that alternative methods such as HAFN \cite{xu2018unsupervised}  tend to deteriorate the overall performance. The possible reasons for the performance drop are: i) the additional hyper-parameters in \cite{xu2018unsupervised,zheng2018ring} require careful manual tuning; ii) the ${l_2}$-norm of the feature embeddings may hurt the generalization capacity and don't cooperate well with IN in early layers. On the contrary, our approach yields consistently better results, validating the efficacy of our FN choice.
	
	\begin{table}[t]
		\centering
		\begin{tabular}{l|c|c|l|l}
			\hline
			Methods & VIPeR & PRID & GRID & i-LIDS \\
			\hline					 
			None & 52.7 & 54.1 & 39.7 & 70.5\\
			+ HAFN \cite{xu2018unsupervised}  & 47.9 & 51.2 & 34.0 & 64.5 \\
			+ IAFN \cite{xu2018unsupervised} & 50.1& 53.8 & 35.6 & 65.8\\
			+ Ring loss \cite{zheng2018ring} & 46.6 & 49.1 & 33.8 & 61.7\\
			+ FN~(Ours) & 53.9 & 60.4 & 41.4 & 74.8\\
			\hline
		\end{tabular}
		\caption{Comparison with other feature norm approaches (cross-domain rank1 accuracy).}
		\label{ab6}
	\end{table}
	
	\keypoint{Compatibility with other backbone networks}
	To  demonstrate the generality of our method, we apply it to the same \reid{} problem using another prevalent CNN architecture, ResNet50 \cite{He2016Deep}. The model is initialized with parameters pre-trained on ImageNet \cite{deng2009imagenet}. The initial learning rate is set to 0.05 for classification layer and 0.005 for other base layers, decayed by 0.1 after 40 epochs. The model is trained for 70 epochs in total. Similar to MobileNetV2, the initial convolution block in ResNet50 is referred to as `Conv1' and the following groups of residual bottlenecks are denoted as `Conv2\_x-Conv5\_x' respectively. By default, for experiments on ResNet50, IN is added to the last block of `Conv2\_x' to `Conv4\_x', with the rest of network unchanged. Also, the FN is applied to the penultimate layer after feature encoding.	

	From Table~\ref{ab2}, the following observations can be made: (1) A consistent improvement over vanilla ResNet50 can be achieved with our {\name} solution, demonstrating its ability to overcome domain-specific biases and enhance the performance of \reid{} DG problems. (2) We evaluate the closely related IBN-a and IBN-b model \cite{pan2018two} for our task, with the network architectures in their original paper. It can be seen that they can greatly boost the cross-domain accuracy, but not as significantly as our {\name} strategy, which  considers and alleviates both style and content domain-biases. (3) As for the design-space analysis on the location and number of IN layers, similar to the result on MobileNetV2 in Table~\ref{ab0}, we can see that more IN in early layers is beneficial, but IN applied to deep layers deteriorates  performance. Note that IN added in `1-3' is exactly the same as ResNet50-IBN-b in \cite{pan2018two}. (4) Comparing `1-4' with `2-4', `1-3' with `2-3', we can see that opposite to the results in Table~\ref{ab0}, instance normalization on Conv1 in ResNet50 has a detrimental effect. (5) Finally, see that  IN and FN together remarkably increase the cross-domain accuracy, confirming their utility in eliminating  statistics disparities across domains and their complementary strength. (6) Though ResNet50 \cite{He2016Deep} gives higher accuracy than MobileNetV2, we still exploit MobileNetV2 as our default backbone network due to its lower number of parameters and flexibility to avoid pre-training on ImageNet \cite{deng2009imagenet}.
	\begin{table}[t]
		\centering
		\begin{tabular}{l|c|c|l|l}
			\hline
			Models & VIPeR & PRID & GRID & i-LIDS \\
			\hline
			ResNet50 \cite{He2016Deep} &  48.5 & 20.3 & 29.0 & 71.3\\
			\hline
			ResNet50-IBN-a \cite{pan2018two} &  52.1 & 39.3 & 36.2 & 70.4\\
			ResNet50-IBN-b \cite{pan2018two} &  54.9 & 48.5 & 40.9 & 68.8\\
			\hline
			ResNet50+IN~(1-2) &  51.6 & 38.2 & 37.2 & 76.5\\		
			ResNet50+IN~(1-3) &  54.9 & 48.5 & 40.9 & 68.8\\	
			ResNet50+IN~(2-3)  &  55.5 & 50.2 & 42.8 & 70.7\\					
			ResNet50+IN~(1-4)  &  55.8 & 48.8 & 36.7 & 73.8\\			
			ResNet50+IN~(1-5)  &  24.6 & 15.5 & 10.6 & 32.8\\
			\hline		
			ResNet50+IN~(2-4)  &  56.5 & 54.9 & 39.2 & 74.4\\						
			ResNet50+FN  &  50.9 & 60.0 & 38.8 & 76.7\\			
			ResNet50+\name & \textbf{59.4} & \textbf{69.6} & \textbf{43.7} & \textbf{78.2}\\ 	 		
			\hline
		\end{tabular}		
		\caption{Cross-domain \reid{} performance and design-space analysis on where to add IN layers using ResNet50 backbone (rank 1 accuracy).}
		\label{ab2}
	\end{table}
	
	\begin{table}[t]
		\centering
		\begin{tabular}{l|l|l|l|l|l}
			\hline
			Dataset & model &rank-1 &rank-5 &rank-10 & mAP \\		
			\hline
			Market-1501  & MobileNetV2 & 77.2  & 89.9  & 93.8 & 53.9\\
			Market-1501  & MobileNetV2+\name & 82.6  & 91.7  & 95.3 & 57.2\\
			
			\hline
			DukeMTMC-reID  & MobileNetV2 & 65.0  & 79.8  & 84.1 & 44.1\\
			DukeMTMC-reID  & MobileNetV2+\name & 71.2  & 82.5  & 86.3 & 48.3\\
			\hline
		\end{tabular}
		
		\caption{Within-dataset results on Market-1501 and DukeMTMC-reID dataset.}
		\label{ab3}
	\end{table}
	
	\keypoint{Within-dataset experiments}
	The previous experiments show that our method significantly improves cross-domain generalisation in \reid{}. Finally we evaluate our method in single-dataset re-ID tasks. The results for popular datasets Market-1501 \cite{zheng2015scalable} and DukeMTMC-reID \cite{zheng2017unlabeled} are shown in Table~\ref{ab3} with the train/test split and protocol in the original papres. We still train the models from scratch with 300 epochs and learning rate starts from 0.01, decayed by 0.1 after 100 epochs. We can see that our method boosts within-dataset performance as well. The increase in rank 1 is 5.4\% and 6.2\% on Market-1501 and DukeMTMC-reID respectively, not as significant as cross-dataset experiments since the bias between train/test within one dataset is much less dramatic. Nevertheless, a performance improvement is still observed, indicating that our strategy can handle intra-domain variations as well.	
	
	\section{Conclusion}
	We proposed a normalization based domain generalization (DG) baseline for person \reid{}. With a combination of Instance and Feature Normalization, style and content biases between domains are alleviated, promoting generalization and transferability of Deep \reid{}. Experiments demonstrate that our approach surpasses contemporary supervised, unsupervised, and purpose-designed DG methods for multi-source cross-domain \reid{}. We believe this fast and simple but effective method provides a strong baseline that will be invaluable to engineers implementing \reid{} in practice as well as a baseline for future research.
	
\keypoint{Acknowledgements} 	This work was supported by China Scholarship Council, EPSRC grant EP/R026173/1, National Natural Science Foundation of China (61772067, 61471032, 61472030) and NVIDIA Corporation GPU donation.
	\bibliographystyle{abbrvnat}
	\bibliography{egbib}
\end{document}